\pdfoutput=1

\documentclass[11pt]{article}

\usepackage{EMNLP2022}

\usepackage{times}
\usepackage{latexsym}

\usepackage[T1]{fontenc}

\usepackage[utf8]{inputenc}

\usepackage{graphicx}
\graphicspath{ {./images/} }

\usepackage{microtype}

\usepackage{inconsolata}
\usepackage{hyperref}

\title{TEDB System Description to a Shared Task on Euphemism Detection 2022}

\author{Peratham Wiriyathammabhum \\
  \texttt{peratham.bkk@gmail.com} \\
  \vspace{-10\baselineskip}
  }

\begin{document}
\maketitle
\begin{abstract}
In this report, we describe our Transformers for euphemism detection baseline (TEDB) submissions to a shared task on euphemism detection 2022. We cast the task of predicting euphemism as text classification. We considered Transformer-based models which are the current state-of-the-art methods for text classification. We explored different training schemes, pretrained models, and model architectures. Our best result of \textit{0.816} F1-score (\textit{0.818} precision and \textit{0.814} recall) consists of a euphemism-detection-finetuned TweetEval/TimeLMs-pretrained RoBERTa model as a feature extractor frontend with a KimCNN classifier backend trained end-to-end using a cosine annealing scheduler. We observed pretrained models on sentiment analysis and offensiveness detection to correlate with more F1-score while pretraining on other tasks, such as sarcasm detection, produces less F1-scores. Also, putting more word vector channels does not improve the performance in our experiments.
\end{abstract}

\section{Introduction}
A shared task on euphemism detection \cite{gavidia2022cats, lee2022searching} is the first installment of a natural language processing (NLP) shared task on a particular figurative language detection, euphemism. Figurative languages, including metaphors, synecdoches, idioms, puns, hyperbole, similes, onomatopoeia, and others, are word uses where the meaning deviates from the literal meaning to convey a complicated, creative and evocative message without directly stating it. In addition, figurative language might use contexts such as relations to other things, actions, social experiences, or images. Figurative languages are ubiquitous since they are filled in countless of our everyday activities without notice \cite{lakoff2008metaphors}. 

Euphemisms are mild or indirect words or phrases being used in place of offensive or unpleasant ones. Moreover, euphemisms are used to mark profanity or politely refer to sensitive and taboo topics such as death, disability, or sickness. The applications of euphemisms involve social interactions such as politics or doctor-patient discourses. Euphemisms can also be dangerous since terrorists can use euphemisms for language manipulation and separate message and meaning \cite{matusitz2016euphemisms}. Also, politely calling terrorism results in semantic deviance and attention away from reality for media and government officials which makes citizens lower their guard while in danger.

\begin{table}[!t] 
\centering
\caption{An Example instance from the shared task dataset. The first sentence is more offensive literally. The phrase ``collateral damage" should be replaced with politeness. The second sentence was revised by using the phrase ``advanced age" to provide more politeness than some possible words or phrases like old, near expiration, or wrinkly.}
\vspace{-0.5\baselineskip}
\label{dtable1}
\begin{tabular}{|l|c|c|}
\hline
Sentence    & Label \\
\hline
 All the deaths were just   & [non-euphemistic] \\
 <collateral damage>  & \\
 in their cause. & \\
 \hline
 In spite of his     & [euphemistic] \\
  <advanced age>,  & \\
   Rollins remains one of     & \\
  jazz's most talented  & \\
  improvisers. & \\
\hline
\end{tabular}
\vspace{-1\baselineskip}
\end{table}

Previous works \cite{gavidia2022cats, lee2022searching} utilize RoBERTa models \cite{liu2019roberta} for sentiment and offensive ratings because politeness is the aim of euphemisms. Euphemisms should make the sentences more positive in sentiment and less offensive \cite{bakhriddionova2021needs}. Our systems build upon these findings and explore transformer-based models which are pretrained for sentiment analysis or offensive detection.

Our best submission ranks $6^{th}$ on the leaderboard. The codes for our systems are open-sourced and available at our GitHub repository\footnote{\url{https://github.com/perathambkk/euphemism_shared_task_emnlp2022}}.

\section{Models}
\subsection{Pretrained Transformers}
Huggingface library \cite{wolf-etal-2020-transformers} is an extensive platform for transformer models \cite{vaswani2017attention}. Huggingface provides many checkpoints for the pretrained transformer suitable to many tasks as a model hub. TweetEval \cite{barbieri2020tweeteval} is a social NLP benchmark where standardized evaluation protocols and strong baselines and employed on seven Twitter classification tasks. The strong baselines later became pretrained model checkpoints loadable via Huggingface.

Diachronic specialization was shown to be lacking in language models \cite{loureiro-etal-2022-timelms} where changes or evolution in time can break current (synchronic - a language at a moment in time without any histories.) language model performances entirely. For example, pre-COVID19 language models will have no knowledge about the pandemic events completely. Diachrony and synchrony are two complimentary viewpoints that were theorized by linguist Ferdinand de Saussure more than a hundred years ago \cite{de2011course}. The paper shares many time-specific language model checkpoints (TimeLMs). 

Specifically, we employed two RoBERTa language model checkpoints from the papers (TweetEval and TimeLMs), one for sentiment analysis (`cardiffnlp/twitter-roberta-base-sentiment-latest') and another for offensiveness detection (`cardiffnlp/twitter-roberta-base-offensive'), as in \cite{gavidia2022cats, lee2022searching}. We finetuned them for euphemism detection as text classification.

\subsection{Convolutional Neural Networks Backend}
Convolutional Neural Networks (CNN) were primarily introduced for visual tasks, firstly, handwritten digit recognition, given its properties in translation invariance for 2D data \cite{lecun1998gradient}. KimCNN \cite{kim-2014-convolutional} proposed a little modification that enables on-top finetuning of CNN over pretrained word vectors for sentence classifications. The results in the paper were from a simple CNN with a little parameter tuning and static vectors. 

We further performed some modifications by concatenating hidden state outputs from all RoBERTa layers as a word vector and instead finetuning the whole model end-to-end. We also used checkpoints from finetuning the pretrained transformers as RoBERTa starting points. We also attempted to combine two word vectors for a multichannel KimCNN and finetuning the model with both word vectors for sentiment analysis and offensiveness detection end-to-end in contrast to freezing one word vector channel as in the original paper.

\begin{table*}[t] 
\centering
\caption{Test F1-scores of different pretrained transformers on euphemism detection. (The number in \textbf{bold} is for the best score, and in \textit{italic} is for the second best.)}
\vspace{-0.5\baselineskip}
\label{stable1}
\begin{tabular}{|l|c|}
\hline
 Pretrained Transformer    & Test F1-score  \\
\hline
`cardiffnlp/xlm-twitter-politics-sentiment' & 0.4693  \\
`Hate-speech-CNERG/dehatebert-mono-english' & 0.6821 \\
`mrm8488/t5-base-finetuned-sarcasm-twitter-classification' & 0.6969 \\
`finiteautomata/bertweet-base-sentiment-analysis' & 0.7349 \\
\hline
a strong finetuned vanilla baseline: `roberta-base' & 0.7776\\
\hline
`sagteam/covid-twitter-xlm-roberta-large' & 0.7776 \\
`cardiffnlp/twitter-roberta-base-offensive' & \textit{0.7838} \\
\hline
another strong finetuned vanilla baseline: `bert-base-cased' & 0.7941 \\
\hline
`cardiffnlp/twitter-roberta-base-sentiment-latest' (TimeLMs) & \textbf{0.8064} \\
\hline
\end{tabular}
\vspace{-0.5\baselineskip}
\end{table*}
\section{Experimental Setup}
\begin{figure}[t]
\begin{center}
   \includegraphics[width=1.00\linewidth]{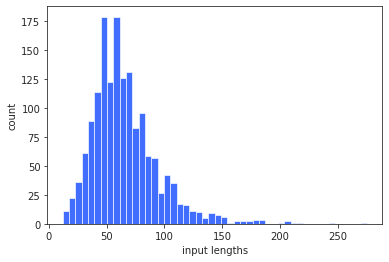}
\end{center}
\vspace{-1.0\baselineskip}
   \caption{\textbf{The distribution of the input length derived from the shared task training set.}}
\label{fig:long}
\label{fig:onecol}
\label{figstat}
\vspace{-1\baselineskip}
\end{figure}

Our input consists of a three-sentence utterance, the sentence before, after, and the sentence containing the euphemistic term. We did not observe any improvements from removing any special characters including the `<' and `>' symbols around the euphemistic term given in the dataset. We used the maximum input length of $150$ tokens since we found that it is the number that fits well as our heuristics with the GPU memory for many reasonable batch sizes ($4 - 20$ in our cases). Also, it seems to cover most data instances given the histogram plotting in Figure \ref{figstat}. We sampled the model at the end of each epoch. The dataset has $1572$ training instances and $393$ test instances. 

All of our experiments were done in the Google Colab setting on NVIDIA Tesla T4 GPUs. We used the batch size in the range of $4-20$ and the learning rate for an AdamW optimizer \cite{loshchilov2018decoupled} in the set of $\{2.5e-5, 2e-5, 1e-5, 7.5e-6\}$ for all experiments. We considered linear annealing scheduler and cosine annealing scheduler with restart. The cycle number is in the set of $\{5, 8\}$. Also, adding a warm-up step does not make any difference so we set the warm-up step to zero in all experiments. 

\subsection{Early Stopping Criterion for Empirical Risk Minimization}
We employed the early stopping with zero patience training strategy schema \cite{prechelt1998early, bengio2012practical}. We varied the training epoch until the training metric saturated with manual monitoring, and then stopped right at the end of that epoch. We tried to split the training data into training and development sets but empirically we found that the data set size is too small to perform accurate estimations/cross-validations on just an efficient held-out schema. For these reasons, we relied solely on our heuristics on the training set instead.

Theoretically simply speaking, given a small data for finetuning, it is not easy to estimate the model performance using a held-out validation set. Leave-one-out cross-validation (LOOCV) is appropriate but might need much more computation costs. Even $k$-fold cross-validation with a high value of $k$, which is a less extreme case of LOOCV, still needs a lot of computation costs. Additionally, if we split a small data, our model might fit the train split, but not the validation split. That model is very unlikely to perform well on the validation split, especially when the training is still underfitting the task, given a small data to train and a data-hungry model with a large capacity. Therefore, it will certainly have a weak upper bound of its error against a model that fits the whole training data. 

This gives us an intuition of training our models just to shatter the whole training data and stop training in a basic train-test setting (empirical risk minimization). In our other simple intuition, it would be weird to withhold some training data from a given small data, implicitly lower the model capacity by (randomly) filtering out some data for an inaccurate generalizability estimation, and let the model predict them wrongly. Also, using more data to train lowers the model variance error term in the bias-variance decomposition framework. 



\begin{table*}[t] 
\centering
\caption{Test F1-scores of different TweetEval pretrained transformers \cite{barbieri2020tweeteval} on euphemism detection. (The number in \textbf{bold} is for the best score, and in \textit{italic} is for the second best.)}
\vspace{-0.5\baselineskip}
\label{stabletweval}
\begin{tabular}{|l|c|}
\hline
 Pretrained Transformer    & Test F1-score  \\
\hline
`cardiffnlp/twitter-roberta-base-stance-climate' & 0.7238 \\
`cardiffnlp/twitter-roberta-base-sentiment' & 0.7238 \\
`cardiffnlp/twitter-roberta-base-stance-feminist' & 0.7306 \\
`cardiffnlp/twitter-roberta-base-stance-abortion' & 0.7446 \\
`cardiffnlp/twitter-roberta-base-emotion' & 0.7588 \\
`cardiffnlp/twitter-roberta-base-emoji' & 0.7615 \\
`cardiffnlp/twitter-roberta-base-stance-hillary' & 0.7651 \\
`cardiffnlp/twitter-roberta-base-hate' & 0.7665 \\
`cardiffnlp/twitter-roberta-base-irony' & \textit{0.7688} \\
`cardiffnlp/twitter-roberta-base-stance-atheism' & \textit{0.7688} \\
\hline
a strong finetuned vanilla baseline: `roberta-base' & 0.7776\\
\hline

`cardiffnlp/twitter-roberta-base-offensive' & \textbf{0.7838} \\
\hline
another strong finetuned vanilla baseline: `bert-base-cased' & 0.7941 \\

\hline
\end{tabular}
\vspace{-0.5\baselineskip}
\end{table*}

\subsection{Finetuning Pretrained Transformers}
We compare many available huggingface hub's pretrained checkpoints we feel suitable for the task, which are multilingual Twitter politics sentiment analysis \cite{antypaspolitics}, hate speech detection \cite{10.1007/978-3-030-67670-4_26}, Twitter sarcasm detection \cite{ghosh2020report, raffel2020exploring}, Twitter English sentiment analysis \cite{nguyen-etal-2020-bertweet, loureiro-etal-2022-timelms}, Multilingual Russian-English Twitter COVID-19 report detection \cite{sboev2021russian}, and offensiveness detection \cite{barbieri2020tweeteval}. The transformer models include BERT \cite{kenton2019bert}, RoBERTa \cite{liu2019roberta}, XLM \cite{conneau-etal-2018-xnli}, XLM-RoBERTa \cite{conneau-etal-2020-unsupervised} and T5 \cite{raffel2020exploring} which are finetuned for the target task and their model parameters are shared on huggingface hub.

From the test F1-scores in Table \ref{stable1}, in which we even report the best result from all model hyperparameter settings in our experiment not reported here for brevity, we tend to confirm the hypothesis in the aforementioned previous works \cite{gavidia2022cats, lee2022searching, bakhriddionova2021needs} which state that euphemism relates with sentiment and offensiveness because the top-2 best scores in the table are sentiment analysis and offensiveness detection. Also, multilingual pretraining seems not to be helpful in this case of English euphemism detection. The `cardiffnlp/twitter-roberta-base-sentiment-latest' RoBERTa-base model seems to outperform the `finiteautomata/bertweet-base-sentiment-analysis' BERTweet model as in the TimeLMs paper \cite{loureiro-etal-2022-timelms} too. Therefore, we further build our models based on these top-2 best scorer pretrained TweetEval/TimeLMs RoBERTa models \cite{gururangan-etal-2020-dont}. We are aware that these top-2 models were among pretrained language models using the most data in TweetEval/TimeLMs.

\subsubsection{TweetEval Pretrained Language Models}
However, when we additionally compared all TweetEval pretrained RoBERTa-base language models finetuned on the euphemism task using our training scheme in Table \ref{stabletweval}, we observed that a TweetEval sentiment analysis model does not perform well at all. Besides, it was pretrained using much less data than the one in TimeLMs (45k vs. 138.86M tweets). Still, in Figure \ref{stable1}, the TimeLMs sentiment classification model performs very well given lots of data. The sentiment classification task might have some correlations with euphemism detection when the model learns well, or just lots of data make it work. 

The best result in Table \ref{stabletweval} is from offensiveness detection with only 11k tweet data. The second best models are irony detection and stance detection in the target domain of atheism. The performances vary based on some degree of euphemisms in the pretrained data. Nevertheless, only the offensiveness detection language model performs better than a finetuned vanilla RoBERTa-base language model. Finally, this is only our evidence-based intuition based on some point estimations of the model performances on euphemism detection. 

We observed high sensitivities in hyperparameter settings in these experiments. Changing some hyperparameters such as patience in early stopping, initial learning rate, learning rate scheduler cycle, or even the random seed can result in significant changes in the results as in typical transformer models which are known to be sensitive to perturbations \cite{dodge2020fine}. Training the `cardiffnlp/twitter-roberta-base-sentiment-latest' model until the training metric is saturated but using a linear scheduler for $10$ epochs instead of the best $15$ epochs and removing special characters can result in $0.6920$ test F1-score, using a linear scheduler for $12$ epochs and removing special characters can result in $0.7301$ test F1-score, which both are significant degradation.

\subsubsection{A Comparison to Vanilla Pretrained Language Models}
\begin{table}[!t] 
\centering
\caption{Test F1-scores of different classifiers on euphemism detection using vanilla pretrained language models. (The number in \textbf{bold} is for the best score.)}
\vspace{-0.5\baselineskip}
\label{stablefeaext}
\begin{tabular}{|l|c|}
\hline
 Model    & RoBERTa-base   \\
\hline
Huggingface's   & 0.5203  \\
classifier &  \\
sklearn logreg & 0.4376  \\
PA classifier & 0.4126 \\
3-NN & \textbf{0.5446} \\
MLP & 0.4545 \\
Decision Tree & 0.4910 \\
Linear SVM & 0.4125 \\
\hline
\end{tabular}
\begin{tabular}{|l|c|}
\hline
 Model    & BERT-base-cased   \\
\hline
Huggingface's   & 0.4197  \\
classifier &  \\
sklearn logreg & 0.5062  \\
PA classifier & \textbf{0.5239} \\
3-NN & 0.4436 \\
MLP & 0.4927 \\
Decision Tree & 0.4315 \\
Linear SVM & 0.4125 \\
\hline
\end{tabular}
\vspace{-1\baselineskip}
\end{table}

\begin{table}[!t] 
\centering
\caption{Validation F1-scores of different classifiers on euphemism detection using vanilla pretrained language models. The split ratio is 0.40. (The number in \textbf{bold} is for the best score.)}
\vspace{-0.5\baselineskip}
\label{stablefeaextval}
\begin{tabular}{|l|c|}
\hline
 Model    & RoBERTa-base   \\
\hline
sklearn logreg & 0.5954  \\
PA classifier & 0.5929 \\
3-NN & 0.6107 \\
MLP & 0.6438 \\
Decision Tree & \textbf{0.6692} \\
Linear SVM & 0.6260 \\
\hline
\end{tabular}
\begin{tabular}{|l|c|}
\hline
 Model    & BERT-base-cased   \\
\hline
sklearn logreg & 0.5929  \\
PA classifier & 0.5700 \\
3-NN & 0.5954 \\
MLP & 0.6056 \\
Decision Tree & \textbf{0.6743} \\
Linear SVM & 0.6031 \\
\hline
\end{tabular}
\vspace{-1\baselineskip}
\end{table}

We additionally conducted experiments on various classifiers using vanilla pretrained language models, like RoBERTa-base and BERT-base-cased, as fixed feature extractors. From Table \ref{stablefeaext} and Table \ref{stablefeaextval}, the validation F1-scores are not good estimations of any test F1-scores. They overestimate all model performances by some large margins of around $0.12\sim0.15$ by their best differences or more. Training a classifier on a fixed feature extractor yields us only at most around $\sim0.54$ test F1-score. This is a large gap compared to the performance of most finetuned language models. Also, the classifier with the best validation score, a decision tree, performs poorly on the test set. We used default parameters for the classifiers and used the same early-stopping training scheme but with an initial learning rate of $2.5e-4$.

\subsection{Finetuning KimCNNs}
We employed the finetuned `cardiffnlp/twitter-roberta-base-sentiment-latest' RoBERTa from the previous subsection for our KimCNN. We used $100$ feature maps and ${3, 4, 5}$ weight length set input. We use a cross-entropy loss function and cosine annealing scheduler for this model type. Other hyperparameters were the same as in the previous subsection.

We got the best result of $0.8158$ test F1-score, approximately $0.01$ improvement over the previous model, simply using a KimCNN backend. However, adding another word vector channel using `cardiffnlp/twitter-roberta-base-offensive’, finetuned in the last subsection, reduces the performance as shown in Table \ref{stable2}. We additionally conducted experiments on removing a large language model and used only static word embeddings. A vanilla KimCNN with either word2vec \cite{mikolov2013distributed} or glove-twitter \cite{pennington-etal-2014-glove}, trained on euphemism detection, works quite well with $0.6807$ and $0.6172$ test F1-scores respectively.

Also, we varied some hyperparameters and observed more stability and faster convergence by simply putting a KimCNN backend on top. The significant degradation in the previous subsection was no longer. The test F1-scores of those models are like $0.8130$ or $0.8132$ which are very close to the best score. We also observed lower scores and slower convergence from using the `cardiffnlp/twitter-roberta-base-sentiment-latest’ directly from the huggingface's hub for KimCNN. So, another pretraining step to the task by finetuning a model from some relevant task helps improve the overall performance. 

\begin{table}[!t] 
\centering
\caption{Test F1-scores of different settings for KimCNNs on euphemism detection. (The number in \textbf{bold} is for the best score.)}
\vspace{-0.5\baselineskip}
\label{stable2}
\begin{tabular}{|l|c|}
\hline
 Model    & Test F1-score  \\
\hline
KimCNNs  & \textbf{0.8158}   \\
 + multichannel & 0.7980 \\
 \hline
KimCNNs & 0.6807 \\
(word2vec) & \\
KimCNNs & 0.6172 \\
(glove-twitter) & \\
\hline
\end{tabular}
\vspace{-1\baselineskip}
\end{table}
\section{Conclusion}
This report describes our baseline systems for a shared task on figurative language processing 2022, euphemism detection. Our best result is from a single-channel KimCNN model using `cardiffnlp/twitter-roberta-base-sentiment-latest', pretrained again for euphemism detection, as a feature extractor. We observed more stability and faster convergence from this training schema. Our results on pretrained transformer models are likely to confirm the previous works \cite{gavidia2022cats, lee2022searching, bakhriddionova2021needs} that euphemism relates with sentiment and offensiveness. Still, we also observed that finetuning a sentiment-based pretrained language model, which pretrained with a rather small dataset, does not perform well. 

\section*{Limitations}
We only sampled a relatively small portion of models and draw conclusions. We also conducted experiments only on one dataset for euphemism detection. We did not perform any strong statistical tests on the models, just point estimations.

The authors are self-affiliated and do not represent any entities. The authors also participated in the shared task under many severe unattended local personal criminal events in their home countries. There might be some unintentional errors and physical limitations based on these unlawful interruptions. Even at the times of drafting this report, the authors suffer from unknown toxin flumes spraying into their places. We want to participate in the shared task because it is fun and educational. We apologize for any errors in this report. We tried our best.

\section*{Acknowledgments}
We would like to thank anonymous reviewers for their constructive feedback, and suggestions for additional experiments. 
\bibliography{anthology,custom}
\bibliographystyle{acl_natbib}

\end{document}